\documentclass[a4paper, 10 pt, conference]{ieeeconf}  % Comment this line out if you need a4paper

\IEEEoverridecommandlockouts                              % This command is only needed if 
\usepackage{graphicx} 
\usepackage{float}
\usepackage[subrefformat=parens,labelformat=parens, caption = false]{subfig}

\usepackage{bm}
\usepackage[linesnumbered,ruled,vlined]{algorithm2e}

\usepackage{amsmath,amssymb,amsfonts}

\usepackage[sorting=none, style=ieee]{biblatex}

\graphicspath{{./Figure/}}
\title{\LARGE \bf
2D Forward Looking Sonar Simulation with Ground Echo Modeling
}

\author{Yusheng Wang$^{1}$, Chujie Wu$^{1}$, Yonghoon Ji$^{2}$, Hiroshi Tsuchiya$^{3}$, Hajime Asama$^{1}$ and Atsushi Yamashita$^{4}$% <-this % stops a space
\thanks{$^{1}$Y.~Wang, C.~Wu, H.~Asama are with the Department of Precision Engineering, Graduate School of Engineering, The University of Tokyo, Japan.~{\tt\small \{wang,chujie,asama\}@robot.t.u-tokyo.ac.jp}}
\thanks{$^{2}$Y. Ji is with the Graduate School for Advanced Science and Technology, JAIST, Japan. {\tt\small ji-y@jaist.ac.jp}}
\thanks{$^{3}$H. Tsuchiya is with the Research Institute, Wakachiku Construction Co., Ltd., Japan. {\tt\small hiroshi.tsuchiya@wakachiku.co.jp}}
\thanks{$^{4}$A.~Yamashita is with the Department of Human and Engineered Environmental Studies, Graduate School of Frontier Sciences, The University of Tokyo, Japan.~{\tt\small yamashita@robot.t.u-tokyo.ac.jp}}
}

\begin{document}

\maketitle
\thispagestyle{empty}
\pagestyle{empty}

%%%%%%%%%%%%%%%%%%%%%%%%%%%%%%%%%%%%%%%%%%%%%%%%%%%%%%%%%%%%%%%%%%%%%%%%%%%%%%%%
\begin{abstract}

Imaging sonar produces clear images in underwater environments, independent of water turbidity and lighting conditions. The next generation 2D forward looking sonars are compact in size and able to generate high-resolution images which facilitate underwater robotics research. Considering the difficulties and expenses of implementing experiments in underwater environments, tremendous work has been focused on sonar image simulation. However, sonar artifacts like multi-path reflection were not sufficiently discussed, which cannot be ignored in water tank environments. In this paper, we focus on the influence of echoes from the flat ground. We propose a method to simulate the ground echo effect physically in acoustic images. We model the multi-bounce situations using the single-bounce framework for computation efficiency. We compare the real image captured in the water tank with the synthetic images to validate the proposed methods. 

\end{abstract}

%%%%%%%%%%%%%%%%%%%%%%%%%%%%%%%%%%%%%%%%%%%%%%%%%%%%%%%%%%%%%%%%%%%%%%%%%%%%%%%%
\section{INTRODUCTION}

Imaging sonars are essential choices for underwater robot perception. Compared to common optical and laser sensors, sonars may not be influenced by water turbidity, refractive distortions, and illumination conditions. A typical imaging sonar, the 2D forward looking sonar (FLS), also known as the acoustic camera, has gained the attention of researchers in this field since it can generate high-quality images in real-time with low power consumption and is compact in size which is suitable for being mounted on consumer class underwater vehicles~\cite{Belcher2002}. It has been applied to various underwater tasks such as tracking, mapping, and navigation \cite{Franchi2021,Wangjoe,Negahdaripour2013,Li2018}. The expenses and the difficulties of underwater experiments are the unavoidable factors hindering the development of the field. Even with complete experiment equipment, it may also need the efforts of divers to ensure the smooth running of the experiment. The means of ground truthing are also limited. For example, it may require constrained experiment settings~\cite{Sung2019} or high-end sensors of great cost \cite{Nielsen2018,Westman2020icra} for accurate sensor motion at a test tank, not to mention the fact that some calibration tasks are still open problems \cite{Li2015}. They lead to the need of building a simulation environment. Simulation of FLS is a non-trivial problem under the spotlight in this field. 

\begin{figure}[tb]
\centering
{\includegraphics[width=1.0\columnwidth]{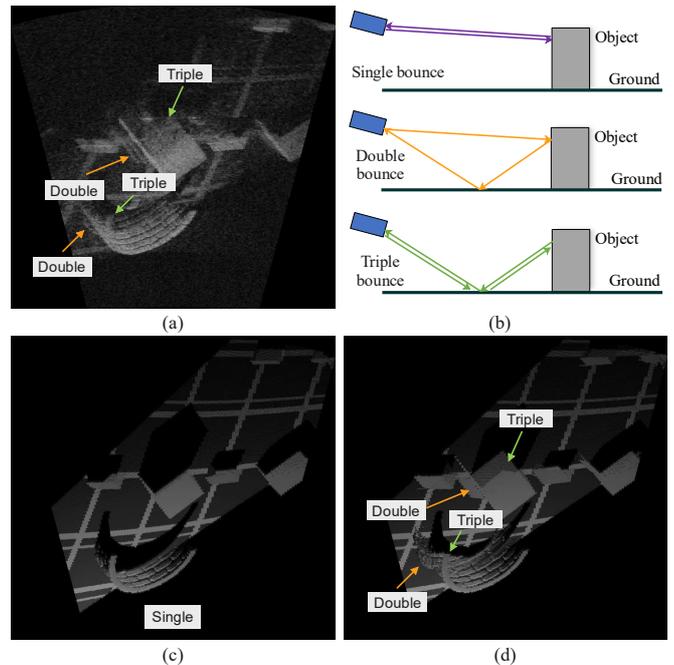}}
\caption{Multi-path reflection problem in acoustic image formation. (a) The real sonar image. (b) Illustration of multiple bounces during sonar imaging. (c) The synthetic sonar image generated only considering the single bounce. (d) The synthetic sonar image generated with the proposed method. Our proposed method can generate realistic images by synthesizing the multi-path phenomenon.} 
\label{fig:catch}
\end{figure}

%Previous 2D forward looking sonar simulators may focus on the geometry model of sonar imaging.
Beginning with binary acoustic image generation \cite{Gu2013}, researchers have tried to model the physical process of sound propagation and calculate the intensities for the pixels \cite{Kwak2015,Mai2018}. The sound is modeled as a large number of rays which may be difficult to process in real time. 
More recent works may focus on accelerating the image generation process using techniques such as GPU computing and rasterization \cite{CERQUEIRA2017,Potokar22icra}. 
Our research group also proposed a differentiable simulator to solve inverse problems like pose refinement by gradient optimization \cite{Wu2023}. However, these works only modeled the single bounce of the signal without considering multi-path reflections. In practice, the double and the triple bounces may significantly influence the image, especially in a water tank as shown in Fig.~\ref{fig:catch}. They may change the geometric information in the acoustic image and eventually degenerate the performance of robot mapping and navigation. Modeling the sonar artifacts like multi-path reflections or reverberation in the acoustic image is a difficult task. Instead, some works proposed neural style transfer methods to generate realistic images from synthetic images by learning the effects from data \cite{Sung2019,Liu2021}. However, it is hard for the network to learn the actual geometric process which may lead to failures during multi-path signal generation for a new scenario. Aykin et al. first modeled the physical phenomenon of ground echoes in \cite{Aykin2016}. This work adopts a similar idea for realistic image generation. 

In this work, we focus on the influence of echoes from flat ground during sonar image simulation. We propose a novel and flexible method to model ground echo in FLS imaging. The ground echo can be categorized into the double bounce and triple bounce as shown in Fig.~\ref{fig:catch}(b). Both situations can be modeled using the single bounce framework by mirroring the sensor or object according to the ground. %The simulation framework can be directly applied to most of the existing rendering engines without low-level code modification and supports GPU acceleration. 
We utilize a dataset captured in a water tank with accurate pose and 3D model ground truth to evaluate our method. It is proved that our proposed method can generate more realistic images by taking ground echoes into consideration. 

\section{Imaging Model}

\begin{figure}[tb]
\centering
\subfloat[ \label{multi}]{\includegraphics[width=0.8\columnwidth]{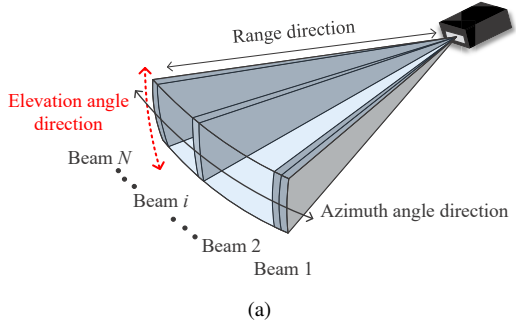}}\\
\subfloat[ \label{proj}]{\includegraphics[width=0.8\columnwidth]{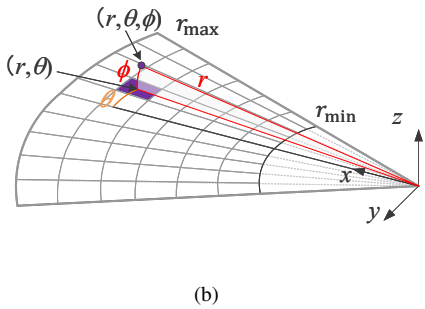}}
\caption{Imaging principle of the 2D forward looking sonar. (a) Multiple beams in azimuth angle direction. (b) The geometry model. } 
\label{fig:projection}
\end{figure}

\subsection{Geometry Model}
The 2D forward looking sonar has $N$ transducers that ensonify multiple beams in azimuth direction as shown in Fig.~\ref{fig:projection}\subref{multi}. A large aperture angle in elevation direction is adopted for imaging. The transducers receive the backscattered time-of-flight signals and process them into an image. For DIDSON-type sonar \cite{Belcher2002}, an acoustic-lens system is used to ensure the signals back to the transducers are from the same directions with emission. Since the sensor only records the time, intensity, and azimuth direction of the signals, the information in the elevation angle direction is missing. 

A 3D point in the sonar coordinate system is usually represented as $(r,\theta,\phi)$ in the polar coordinate system, which can be transformed into the Euclidean coordinate as follows. 

\begin{equation}
	\begin{bmatrix}
	X_c \\ Y_c \\ Z_c 
	\end{bmatrix}
	=
	\begin{bmatrix}
	r\cos  \phi \cos \theta \\ r \cos \phi \sin \theta \\ r \sin \phi 
	\end{bmatrix}.
\end{equation}

The corresponding position in the acoustic image is $(r,\theta)$ which can be equivalently seen as a projection to the $\phi=0$ plane. The raw image is a matrix where each row refers to the range and each column refers to the azimuth angle. It can be transferred into Euclidean coordinates as follows. 

\begin{equation}
	\begin{bmatrix}
	x_c & y_c 
	\end{bmatrix}^{\top}
	=
	\begin{bmatrix}
	r \cos \theta & r  \sin \theta 
	\end{bmatrix}^{\top}.
\end{equation}

\subsection{Image Formation}

\begin{figure}[tb]
\centering
\includegraphics[width=0.8\columnwidth]{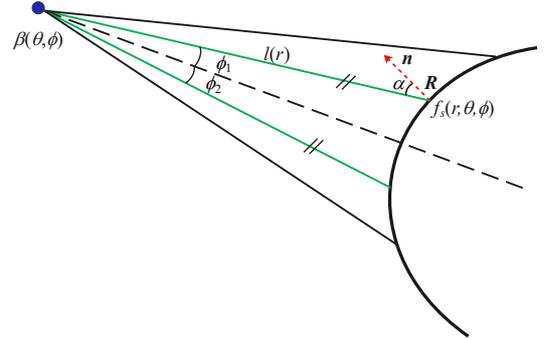}
\caption{A beam slice during sonar imaging. The beam is separated into rays. The backscattered signal is used to generate the image. } 
\label{fig:backscatter}
\end{figure}

To model the ultrasound propagation process, we consider one beam from one transducer as shown in Fig.~\ref{fig:backscatter}. Separating the beam into rays, we do not discuss multipath reflections in this section, so the backscattered signal is from the direction of emission. The strength of the source is a function of $(\theta,\phi)$, written as $\beta(\theta,\phi)$. The ray from the source travels in the medium and hits the object. The ratio of attenuation $l(r)$ is related to the range. The ratio of energy loss on the object surface can be modeled in terms of bidirectional reflectance distribution function (BRDF) as $f_s(r,\theta,\phi)$. 
Denoting the $I_a(r,\theta)$ refers to pixel intensity at $(r,\theta)$, the rendering equation is described as follows. 
\begin{equation}
    I_a(r,\theta) = \int_{\phi_{\rm min}}^{\phi_{\rm max}}\beta(\theta,\phi)l(r)f_s(r,\theta,\phi)d{\phi},
\end{equation}

For one beam, there may be multiple rays with the same range but different $\phi$ angles as in Fig.~\ref{fig:backscatter}, such signal integrates during image formation. 

\section{Simulation Framework}

\begin{figure}[tb]
\centering
\includegraphics[width=1.0\columnwidth]{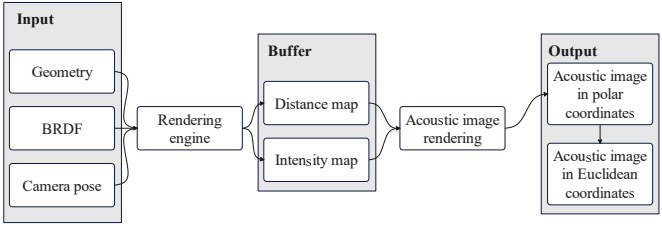}
\caption{The single-bounce simulation framework. The intermediate forms, distance map and intensity map, can be generated using the existing rendering engines. Acoustic images can be calculated from the distance and intensity maps.} 
\label{fig:framework}
\end{figure}

\begin{figure}[tb]
\centering
\includegraphics[width=1.0\columnwidth]{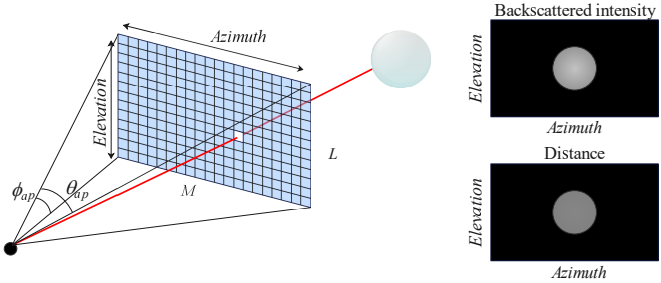}
\caption{Explanation of the sampling process. The backscattered signal and the corresponding distance can be stored in two matrices.} 
\label{fig:sampling}
\end{figure}

This section introduces the backbone of the single-bounce simulation framework. Inspired by \cite{CERQUEIRA2017}, this work also uses rays to sample the scene and records the 3D and the backscattered signal information in matrices in the perspective camera view. As shown in Fig.~\ref{fig:framework}, we first set the scene parameters such as the geometry of the scene, the material of the object, and the camera pose. Then, ray tracing is used to generate distance and intensity buffers. The process is shown in Fig.~\ref{fig:backscatter}. In total, $M\times L$ rays are generated to sample the scene. The distances between the object surface and the sensor center are recorded in a $M\times L$ matrix $D_f$. On the other hand, the backscattered signals of the $M\times L$ rays will be recorded in another $M\times L$ matrix $I_f$. After sampling the scene, it requires integration along the elevation angle as in Eq.~(3). The implementation is described in Algorithm~\ref{algo:sj1} \cite{Wangicra2021}. 

\begin{algorithm} 
\DontPrintSemicolon 
%\KwIn{$\textbf Z_1$, $\textbf S$, $\textbf U$, $\textbf x$, $\textbf y$ } 
\KwIn{$I_f$, $D_f$ } 
\KwOut{$I_a$} 
Initialize $I_a$ using zero matrix $O_{m \times l}$\;
\For{$p = 0$ to $m-1$}{\For{$q = 0$ to $n-1$}{
   $r \xleftarrow{} D_f(p,q) $, $i \xleftarrow{} I_f(p,q) $\;
   $d=\lfloor(r-r_{\rm min})/r_{\rm res}\rfloor$\;
   \If{$0\leq d<l$}{
   $I_a(p,d) = I_a(p,d) + i$\;}}}
\KwRet{$I_a$}\;
\SetAlgoRefName{1} 
\caption{Acoustic Image Formation Function} 
\label{algo:sj1}
\end{algorithm}

This work simplifies the ultrasound propagation model by assuming the beam pattern is uniform, the object surface reflection is Lambertian. The loss during transmission is based on the inverse square law. In practice, there will usually be a time-variant gain (TVG) to compensate for the energy loss during transmission in the medium. The terms in Eq.~(3) can be modeled as follows.

\begin{equation}
    \beta(r,\theta) = C,
\end{equation}
\begin{equation}
    l(r) =\left \{
    \begin{array}{lcl}
      1 &   & \text{w TVG}  \\
     \frac{1}{r^2}  &   & \text{w/o TVG}
    \end{array}.
    \right.
\end{equation}
\begin{equation}
    f_s = \rho\cos\alpha,
\end{equation}
where $\rho$ refers to the diffuse albedo of the surface ranges from 0 to 1 and $\alpha$ is the incidence angle. For objects like concretes with a coarse surface, the $\rho$ value is set higher to 0.8. And for objects like smooth plastic, the value is set lower to 0.4. It is vital to mention that the above terms can be changed to more complex models by modifying the setting in rendering engines. 

\section{Ground Echo Modeling}

\begin{figure}[tb]
\centering
\subfloat[ \label{case1}Case 1]{\includegraphics[width=0.475\columnwidth]{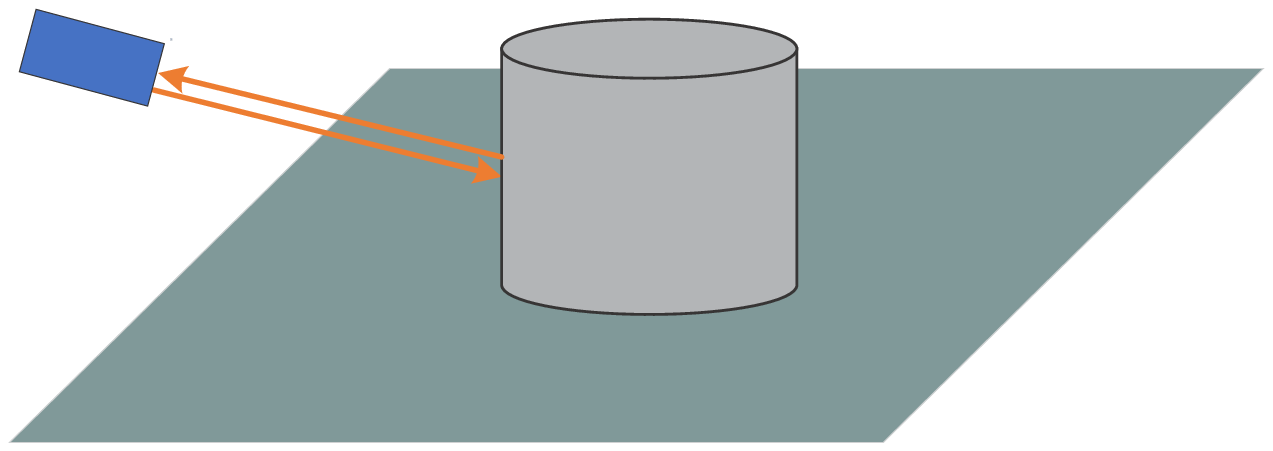}}\enskip
\subfloat[ \label{case2}Case 2]{\includegraphics[width=0.475\columnwidth]{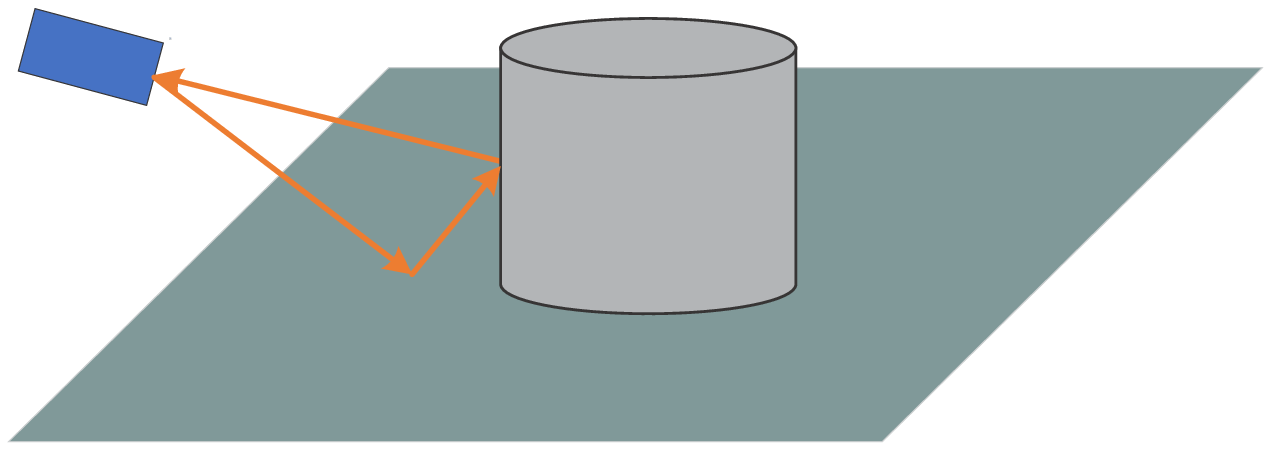}}\\
\subfloat[ \label{case3}Case 3]{\includegraphics[width=0.475\columnwidth]{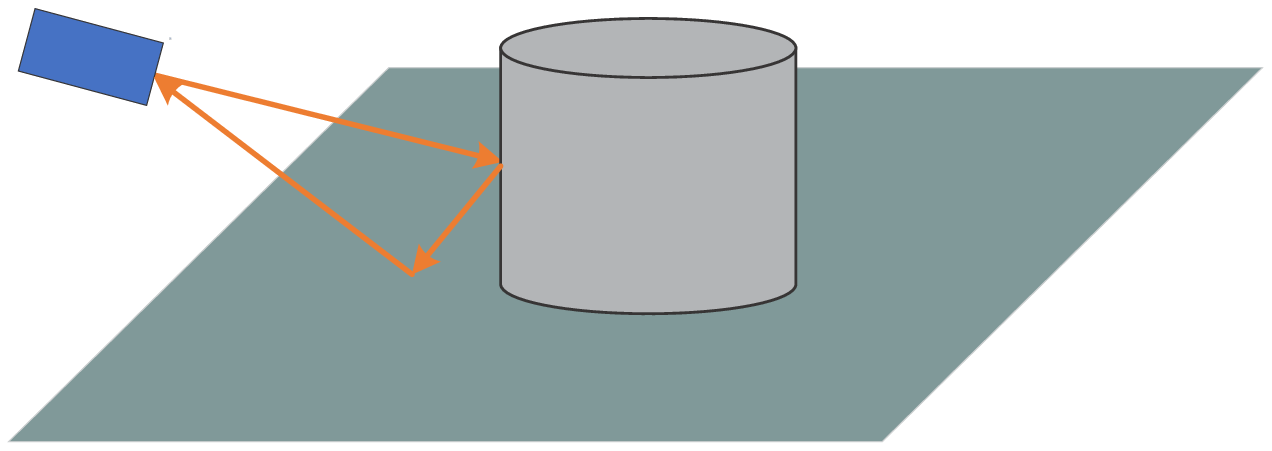}}\enskip
\subfloat[ \label{case4}Case 4]{\includegraphics[width=0.475\columnwidth]{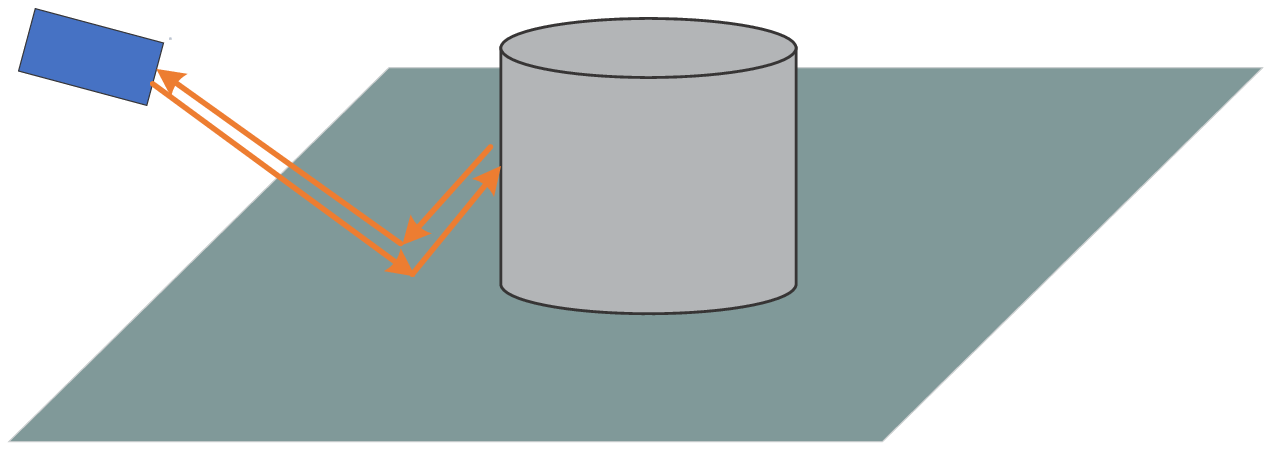}}\\
\caption{Multi-path reflection caused by the ground echo. (a) shows the single bounce case. (b) and (c) are the double bounce cases. (d) is the triple bounce case.  } 
\label{fig:groundecho}
\end{figure}

This work discusses a typical case of the multi-path problem: the influence caused by the ground echo. This frequently happens when collecting data in a water tank with the ground using concrete, metal, ceramic, or plastic pallets. The reflection model of the ground can be described as the sum of specular and diffuse components. The specular components in the previous works are usually not considered. Empirically, the secondary reflection of diffuse components can be ignored but the specular components may be too strong to be ignored, depending on the material types. As shown in Fig.~\ref{fig:groundecho}, signals with more than one bounce time may be reflected back to the transducers. This may significantly change the appearance of the image. It is worth mentioning that we only consider the diffuse components of the objects on the ground.

The paths of major signals received by the sensor are shown in Fig.~\ref{fig:groundecho}. Here case 1 in Fig.~\ref{fig:groundecho}\subref{case1} shows the backscattering case which is the primary signal received. In Fig.~\ref{fig:groundecho}\subref{case2}, the emitted ray first hits the ground and is reflected in a specular reflection manner, denoted as case 2. Then, it hits the object and after diffuse reflection, the signal returns to the sensor. On the other hand, the ray in case 3 (Fig.~\ref{fig:groundecho}\subref{case3}) first hits the object, then reflected on the ground, and is received by the sensor at last. Cases 2 and 3 can be categorized as the same situation. The situation in case 4 as shown in Fig.~\ref{fig:groundecho}\subref{case4} is different, the ray bounces 3 times during propagation, which is twice on the ground and once on the object's surface. 

\begin{figure}[tb]
\centering
\subfloat[ \label{mirror1}]{\includegraphics[width=0.475\columnwidth]{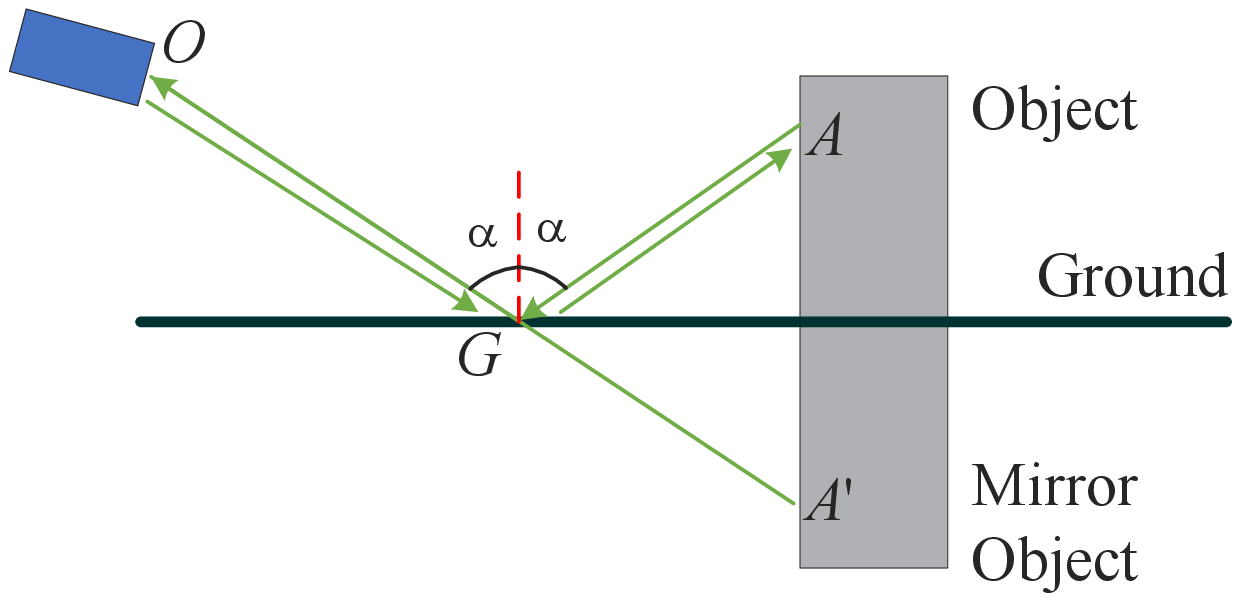}}\enskip
\subfloat[ \label{mirror2}]{\includegraphics[width=0.475\columnwidth]{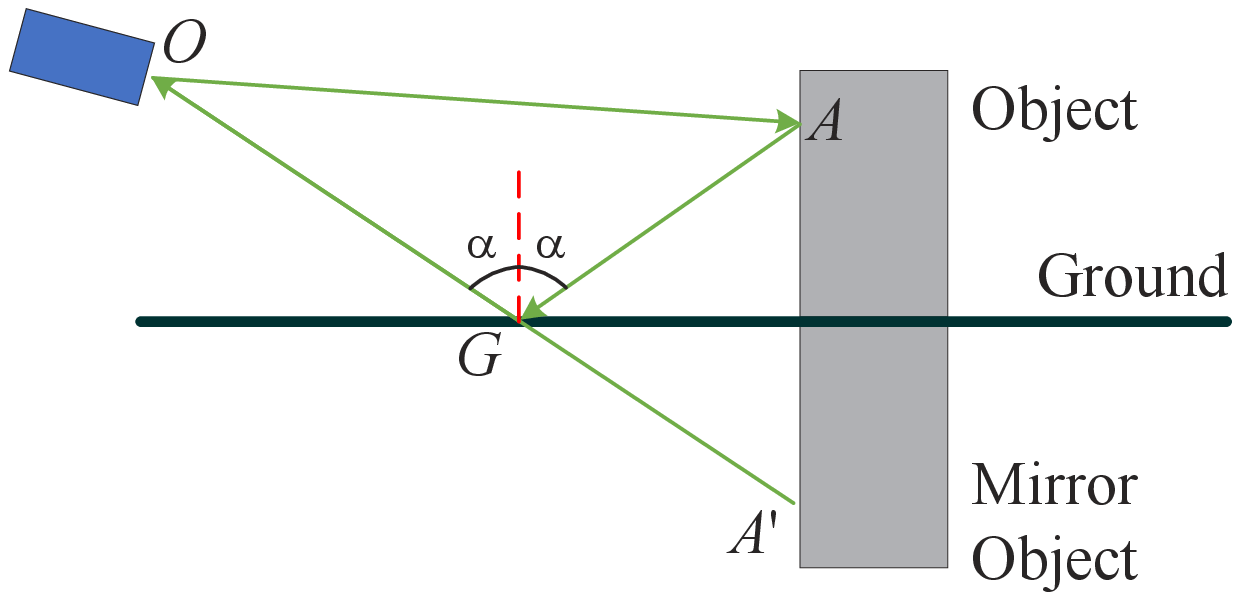}}\\
\subfloat[ \label{mirror3}]{\includegraphics[width=0.475\columnwidth]{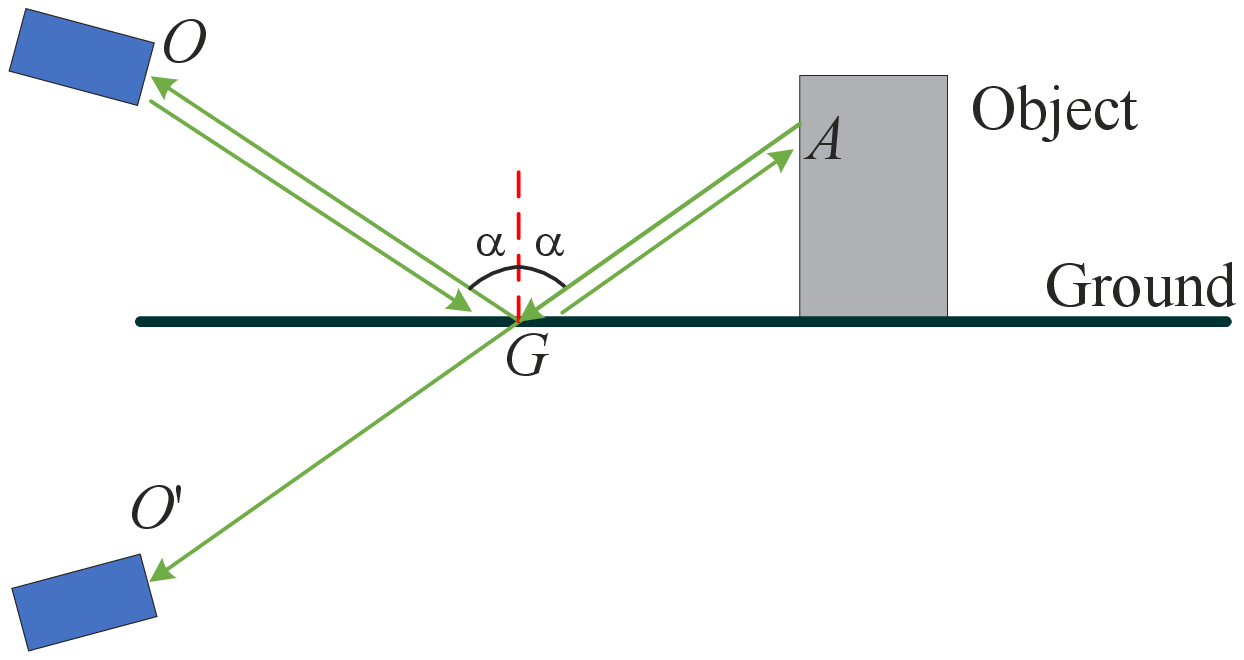}}\enskip
\subfloat[ \label{mirror4}]{\includegraphics[width=0.475\columnwidth]{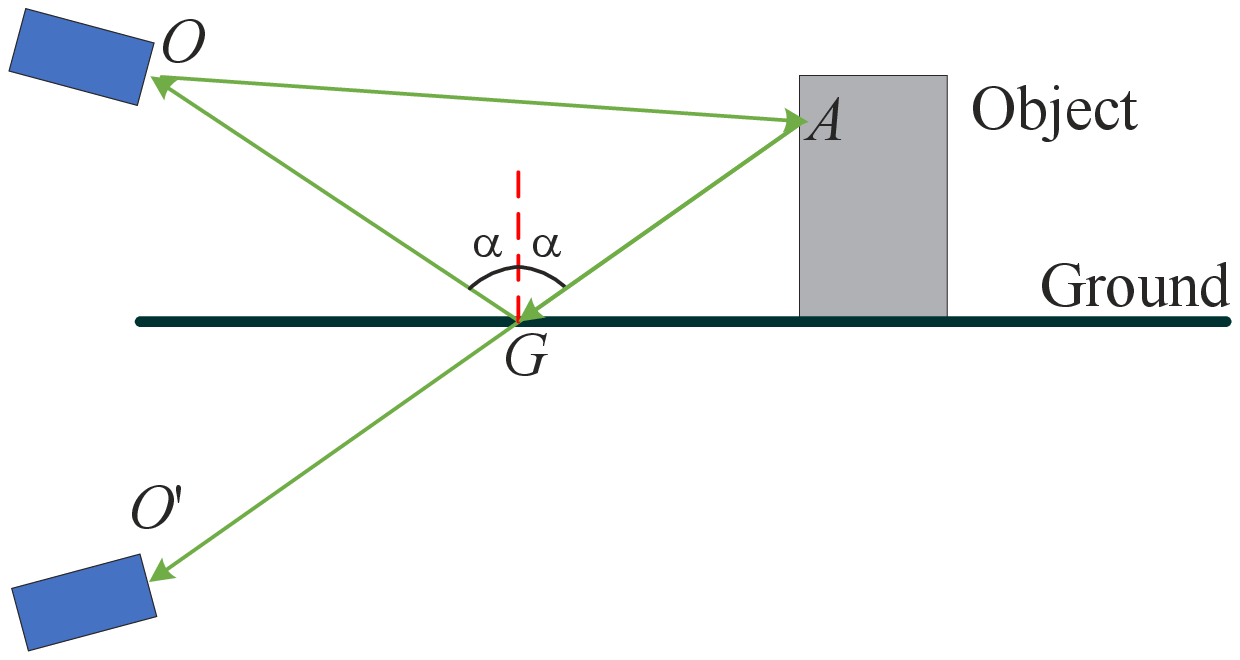}}\\
\caption{Modeling of multi-path reflection caused by the ground echo. (a) Mirroring the object to simulate case 4. (b) Mirroring the object to simulate cases 2 and 3. (c) Mirroring the sensor to simulate case 4. (d) Mirroring the sensor to simulate cases 2 and 3. } 
\label{fig:groundechomirror}
\end{figure}

One of our aims is to reflect the aforementioned phenomena into the simulator. We want to modify the framework in Section~III to generate synthetic images with ground echo effects. For non-direct signals, since we only consider the specular components of the ground, it is possible to model them as shown in Fig.~\ref{fig:groundechomirror} by mirroring the object or camera.

\begin{figure}[tb]
\centering
{\includegraphics[width=0.9\columnwidth]{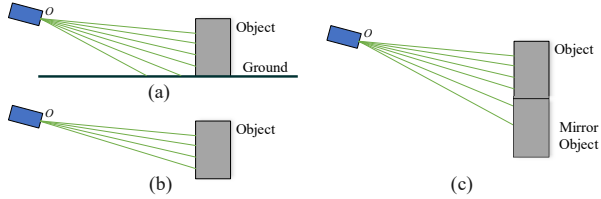}}
\caption{Rendering of the scenes to generate synthetic images. We trace the single-bounce rays of three types of scenes to model the ground echoes. (a) Rendering scene $S_1$ with both object and ground. (b) Rendering scene $S_2$ with the object only. (c) Rendering scene $S_3$ with the object and the mirror object.} 
\label{fig:echoprocess}
\end{figure}

Figures~\ref{fig:groundechomirror}\subref{mirror1} models case 4. Ray path $OG$, $GA$ can be equivalently seen as path $OA'$. The synthetic image can be achieved by rendering the scene $S_1$ in Fig.~\ref{fig:echoprocess}(a) first without considering multiple bounces, denoting the output image as $I_a^{og}$. Then, we only render the object as scene $S_2$ and output the result $I_a^o$. The direct signal from the ground can be represented as follows.
\begin{equation}
I_a^g=I_a^{og}-I_a^o.
\end{equation}

Finally, we render the scene $S_3$ in Fig~\ref{fig:echoprocess}(c), denoting as $I_a^{oo'}$. The signal caused by multi-path reflection in case 4 can be written as follows. 

\begin{equation}
I_a^{o'} = I_a^{oo'}-I_a^o.  
\label{eq:case4}
\end{equation}

For case 4, the final result is as follows. 
\begin{equation}
I_a^{c4fin} = I_a^{o'}+I_a^o+I_a^g = I_a^{oo'}+I_a^g.  
\end{equation}

In fact, mirroring the camera as in Fig.~\ref{fig:groundechomirror}\subref{mirror3} will generate the same result. This work mirrors the object as an implementation. 

For cases 2 and 3, the problem is more complex. The total ray path is $OA+AG+GA$ as shown in  Fig.~\ref{fig:groundechomirror}\subref{mirror2}. Since $AG+GO=OA'$, the reflected intensity will be recorded at position $\frac{OA'+OA}{2}$ in the range direction. It is necessary to compute with the help of distance maps. We first generate the distance map $D_f^{oo'}$ for the scene in Fig~\ref{fig:echoprocess}(c). The multi-path reflection backscattered intensity map can be represented as 
\begin{equation}
I_f^{o'}=I_f^{oo'}-I_f^o,
\label{eq:case23}
\end{equation}
then the next step is to find the distance traveled for the non-zero intensity signals. $OA'$ can be found by looking up to $D_f^{oo'}$. For $OA$, it is necessary to mirror the 3D point $A'$ to $A$ and compute the distance $OA$ in the world coordinates. At last, we add the intensity caused by $A$ to the position $\frac{OA'+OA}{2}$ in the acoustic image. The algorithm for adding ground echo effect is as follows. Here, the 2D position in $D_f$ and $I_f$ corresponds to $A'$ is $(p,q)$. 

\begin{algorithm} 
\DontPrintSemicolon 
%\KwIn{$\textbf Z_1$, $\textbf S$, $\textbf U$, $\textbf x$, $\textbf y$ } 
%\KwIn{$I_f$, $D_f$ } 
%\KwOut{$I_a$} 
Render $I_f^{og}$, $D_f^{og}$, $I_f^{o}$, $D_f^{o}$, $I_f^{oo'}$, $D_f^{oo'}$\;
Compute $I_a^{og}$ from $I_f^{og}$, $D_f^{og}$, $I_a^o$ from $I_f^{o}$, $D_f^{o}$, $I_a^{oo'}$ from $I_f^{oo'}$, $D_f^{oo'}$\;
// Case 4 components\;
Compute $I_a^{o'}$ based on Eq.~(\ref{eq:case4})\;
// Cases 2 and 3 components\;
Initialize $I_a^{c23}$ using zero matrix $O_{m \times l}$\;
Compute $I_f^{o'}$ based on Eq.~(\ref{eq:case23})\;
\For{$p = 0$ to $m-1$}{\For{$q = 0$ to $n-1$}{
\If{$I_f^{o'}(p,q)>0$}{
   $d_{OA'}\xleftarrow{}D_f^{oo'}(p,q)$\;}
   Compute the 3D position of $A'$ based on $(p,q,d_{OA'})$\;
   Mirror $A'$ to $A$ in world coordinates, compute $d_{OA}$\;
   \If{$A$ is within in sensor scope}{
    $d = (d_{OA}+d_{OA'})/2$.\;
    $I_a^{c23}(p,d)=I_a^{c23}(p,d)+I_f^{o'}(p,q)$\;
   }
   }}
// Integrate secondary reflection components with direct reflection\;
$I_a=I_a^{og}+I_a^{c23}+I_a^{o'}$\;
\KwRet{$I_a$}\;
\SetAlgoRefName{2} 
\caption{Ground Echo} 
\label{algo:sj2}
\end{algorithm}

The multi-path reflections can be calculated by sampling single-bounce rays from the three scenes in Fig.~\ref{fig:echoprocess}. In other words, it is not necessary to trace rays up to three bounce times, which saves computational costs.  

\section{Experiment}
\subsection{Dataset}

\begin{figure}[tb]
\centering
\subfloat[ \label{object}]{\includegraphics[width=0.8\columnwidth]{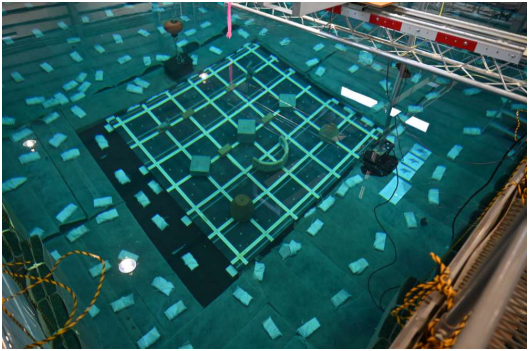}}\\
\subfloat[ \label{device}]{\includegraphics[width=0.8\columnwidth]{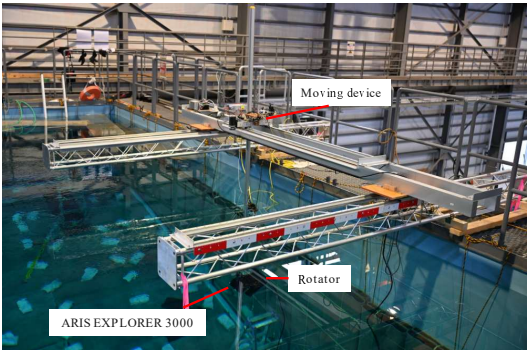}}
\caption{Water tank. } 
\label{fig:tank}
\end{figure}

To evaluate the proposed method, we utilized a dataset collected in a large-scale water tank \cite{Wang2022iros}. We used FLS ARIS EXPLORER 3000 We first collected the ground truth 3D information of the objects in the land environment and then set them up in the water tank by the diver. We measured the relative pose between the scene and the FLS using rulers. However, since the FLS center cannot be accurately measured physically, we also carried out an extrinsic calibration using the acoustic image \cite{Wang2022oceans}. The basic idea is to compare the synthetic image with the real image with the relative pose in the edge domain. If the pose is correct, the difference between the two edge images should be minimized. After extrinsic calibration, the other poses between the scene and the FLS were given by the control input of the moving device and the rotator. We rotated the FLS with roll motion (i.e., around the $x$ axis) at 51 positions. In total, 1,648 images were collected. With known relative pose and the 3D model, it is possible to generate real and synthetic pairs for evaluation of the proposed method.  
\subsection{Results}

\begin{figure*}[tb]
\centering
\subfloat[ \label{r1}]{\includegraphics[width=0.475\columnwidth]{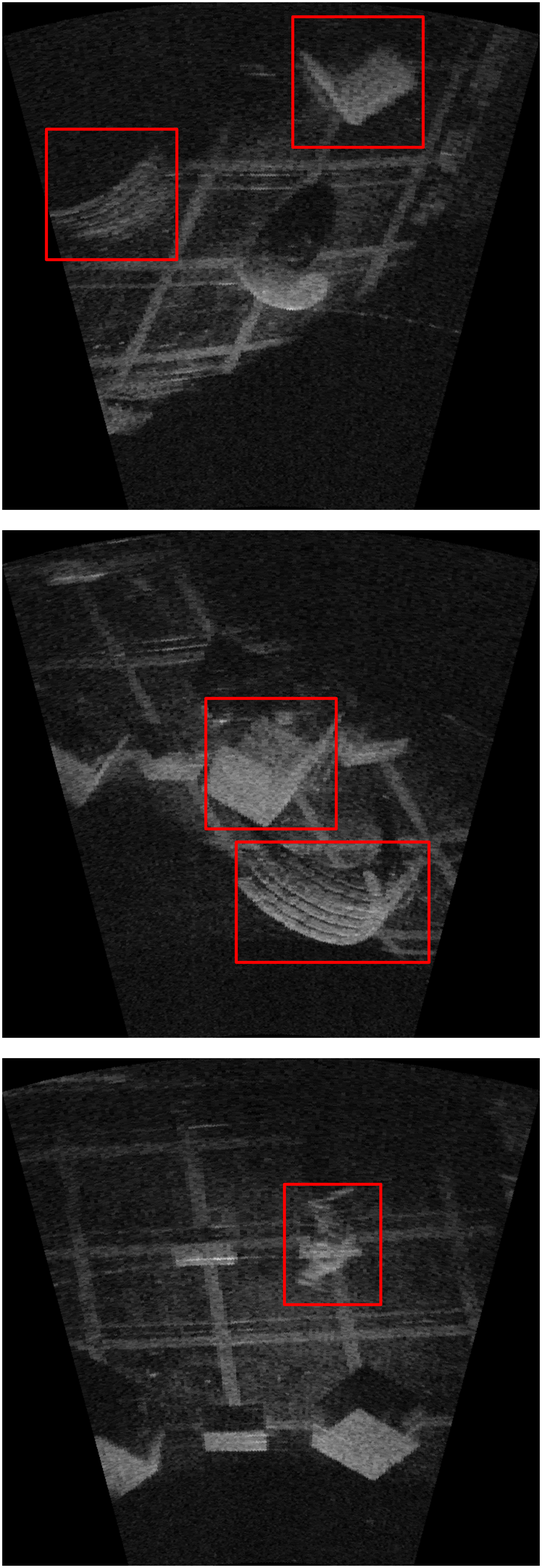}}\enskip
\subfloat[ \label{r2}]{\includegraphics[width=0.475\columnwidth]{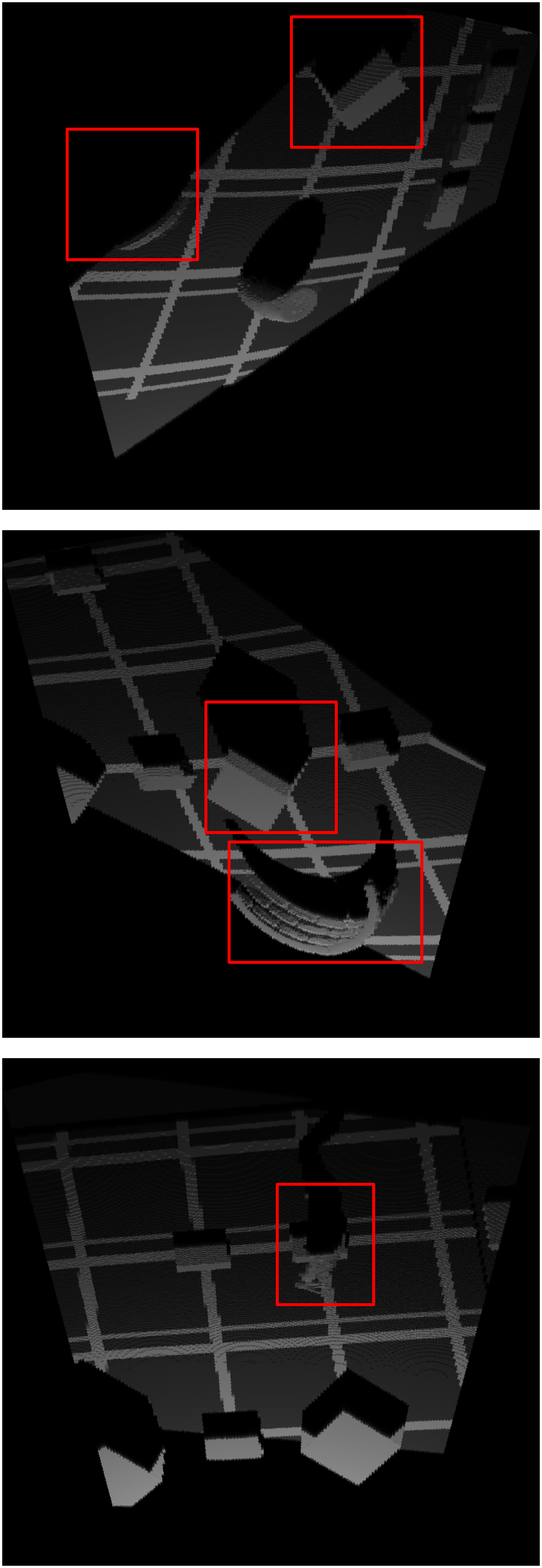}}\enskip
\subfloat[ \label{r3}]{\includegraphics[width=0.475\columnwidth]{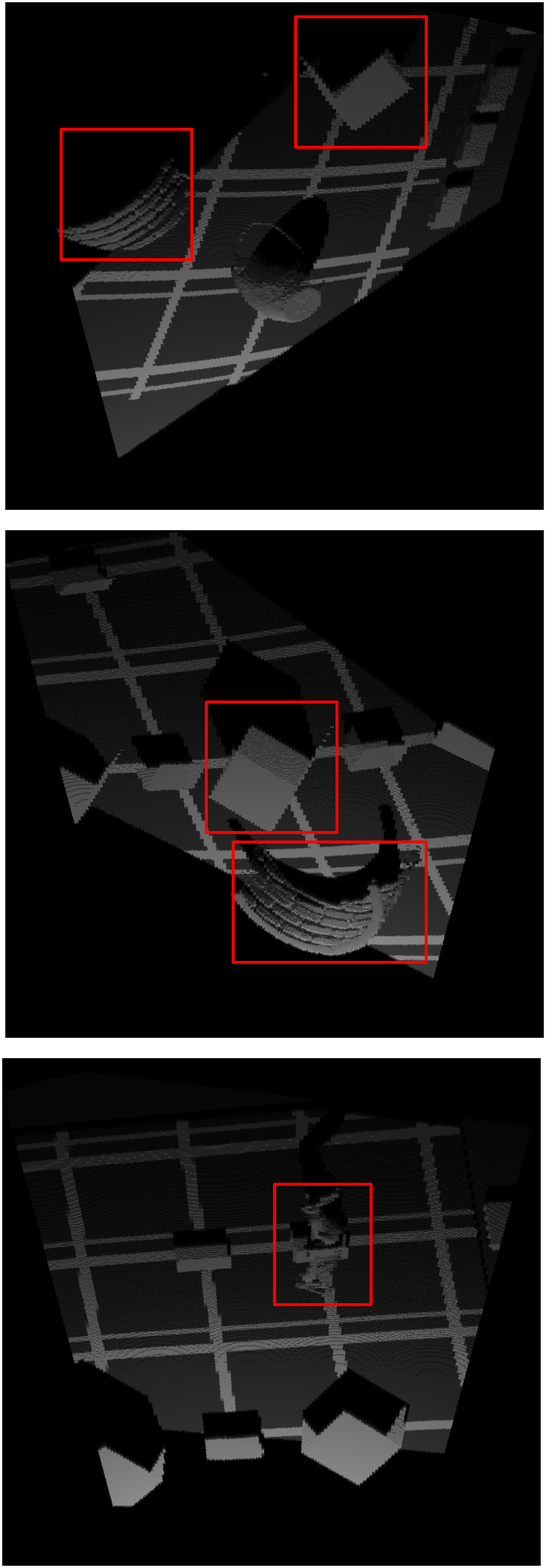}}\enskip
\subfloat[ \label{r4}]{\includegraphics[width=0.475\columnwidth]{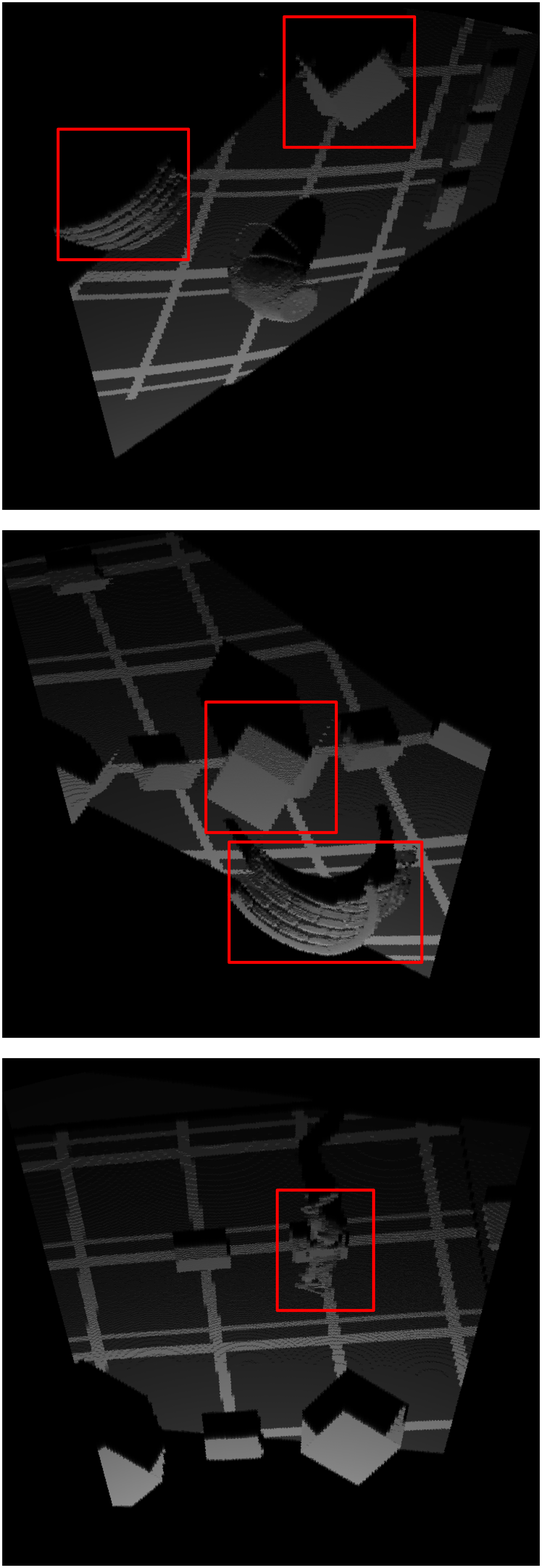}}
\caption{Examples of acoustic image simulation with ground echo modeling. (a) Real images captured in a water tank. (b) Synthetic images from single bounce. (c) Synthetic images from single and triple bounces. (d) Synthetic images with three types of bounces.  } 
\label{fig:result}
\end{figure*}

\begin{table}[tb]
	\centering
	\caption{Quantitative Results}\label{tab:1}
	\scalebox{1.0}{
		\begin{tabular}{|c|c|c|}
			\hline
			&  PSNR & MSE\\ 
			\hline
			Single-GT (Polar) & 17.90  & 104.67 \\
			\hline
			Single and triple-GT (Polar) & 18.44  &  104.38\\
                \hline
			Proposed-GT (Polar) & \bf{18.48}  &  \bf{104.33}\\
			\hline
			Single-GT (Fanshape)& 19.15   & 79.35 \\
			\hline
			Single and triple-GT (Fanshape) & 19.72  &  79.10\\
                \hline
			Proposed-GT  (Fanshape) & \bf{19.76}   &  \bf{79.06}\\
               \hline
		\end{tabular}
	}
\end{table}
We evaluated the results qualitatively and quantitatively. Table~I shows the quantitative results. We used peak signal-to-noise ratio (PSNR) and mean square error (MSE) as the metrics, noting that the image is normalized to 0$\sim$255 for MSE. For PSNR, the results are the higher the better, and for MSE, it is the opposite. We evaluated both polar coordinates images and Euclidean coordinates images (Fanshape). With triple bounce modeling, PSNR increased by +0.54 dB and +0.57 dB in polar coordinates and Euclidean coordinates, respectively. The PSNR can be further increased by +0.04~dB for images in both coordinates by modeling double bounce. However, the influence of double bounce is not as obvious as triple bounce. This is because triple bounce usually exists at shadow regions and influences a large number of pixels. It is worth mentioning that noise simulation was not included in this work, which may decrease the PSNR. Other aspects, such as the overall brightness and contrast may also influence the results. However, this work focused more on geometry. It can be known that our work generates synthetic images with better geometry consistency compared to the real images, as shown in Fig.~\ref{fig:result}. Figure~\ref{fig:result}(a) shows the real image, the red bounding boxes emphasize the region to focus. Figure~\ref{fig:result}(b) illustrates the synthetic image with single bounce modeling only. By considering triple bounce as in Fig~\ref{fig:result}(c), more realistic images can be generated. It can also tell that the double bounce components only exist in a small number of pixels and may behave like noise as shown in Fig.~\ref{fig:result}(d). Currently, it takes about 42 seconds to render an image (128$\times$1288) considering all the bounces. The Python code for acoustic image imaging requires optimization for real-time performance.   
\section{CONCLUSIONS}
In this work, a novel method was proposed to model the ground echo in 2D acoustic images for simulation. By modeling double and triple bounces, more realistic images can be generated. The triple-bounce component may significantly influence the geometry information, which should be paid attention to for further tasks like 3D reconstruction. 

Future work may include decomposing the components with different bounce times and generating better 3D reconstruction results by taking the ground echo into consideration. It is also important to optimize the code and framework for real-time performance. 


%%%%%%%%%%%%%%%%%%%%%%%%%%%%%%%%%%%%%%%%%%%%%%%%%%%%%%%%%%%%%%%%%%%%%%%%%%%%%%%%

%%%%%%%%%%%%%%%%%%%%%%%%%%%%%%%%%%%%%%%%%%%%%%%%%%%%%%%%%%%%%%%%%%%%%%%%%%%%%%%%

%%%%%%%%%%%%%%%%%%%%%%%%%%%%%%%%%%%%%%%%%%%%%%%%%%%%%%%%%%%%%%%%%%%%%%%%%%%%%%%%

%%%%%%%%%%%%%%%%%%%%%%%%%%%%%%%%%%%%%%%%%%%%%%%%%%%%%%%%%%%%%%%%%%%%%%%%%%%%%%%%

%\bibliography{MainBIB}
\printbibliography

\end{document}